\documentclass[sn-mathphys-ay]{sn-jnl}

\usepackage{graphicx}%
\usepackage{multirow}%
\usepackage{amsmath,amssymb,amsfonts}%
\usepackage{amsthm}%
\usepackage{mathrsfs}%
\usepackage[title]{appendix}%
\usepackage{xcolor}%
\usepackage{textcomp}%
\usepackage{manyfoot}%
\usepackage{booktabs}%
\usepackage{algorithm}%
\usepackage{algorithmicx}%
\usepackage{algpseudocode}%
\usepackage{listings}%
\usepackage{colortbl}

\usepackage{enumitem}
\usepackage{pifont}
\usepackage{hyperref}
\usepackage{bbm}

\theoremstyle{thmstyleone}%

\theoremstyle{thmstyletwo}%

\newtheorem{theorem}{Theorem}
\newtheorem{definition}{Definition}
\newtheorem{example}{Example}

\newcommand{\IR}{\mathrm{IR}}
\newcommand{\TPR}{\mathrm{TPR}}        
\newcommand{\FPR}{\mathrm{FPR}}        
\newcommand{\TNR}{\mathrm{TNR}}        
\newcommand{\TP}{\mathrm{TP}}        
\newcommand{\FN}{\mathrm{FN}}        
\newcommand{\FP}{\mathrm{FP}}        
\newcommand{\TN}{\mathrm{TN}}        
\newcommand{\mcc}{\mathrm{MCC}}

\theoremstyle{thmstylethree}%

\begin{document}

\title[The Matthews Correlation Coefficient: Fifty years]{The Matthews Correlation Coefficient: Fifty years}


\author{\fnm{Jesús S.} \sur{Aguilar--Ruiz}}\email{aguilar@upo.es}



\affil{\orgdiv{School of Engineering}, \orgname{Pablo de Olavide University}, \orgaddress{\city{Seville}, \postcode{41013}, \country{Spain}}}




\abstract{Fifty years after its introduction, the Matthews Correlation Coefficient ($\mcc$) remains central---and debated---in binary classification. This survey offers a comprehensive retrospective and an in-depth analysis of $\mcc$, covering its historical development, mathematical foundations, and implications for practice. We clarify long-standing misconceptions, formalize its equivalences (to Pearson’s correlation and to $\chi^2$ in 2×2 tables), and examine its behavior under class imbalance. Beyond $\mcc$, we develop analytical (derivatives) and graphical (contour plots) frameworks that generalize to any confusion-matrix-based metric. These frameworks expose structural properties---bias, robustness, sensitivity---that scalar summaries conceal, and they provide actionable guidance for selecting and interpreting performance measures in applications where reliability and fairness matter. Together, our results position $\mcc$ within a broader toolkit while clarifying when it is informative---and when alternatives may be preferable.}




\maketitle

\section{Introduction}\label{sec:introduction}

The ability to generate reliable forecasts has become essential in supporting decision-making, especially in contexts where reliability is critical. Models designed for this purpose have demonstrated their value across various fields and are often known as predictive systems \citep{Gent2013}, classifier systems \citep{Ho1994}, or diagnostic systems \citep{Swets1998}.

The development of predictive models is largely driven by a broad spectrum of machine learning techniques, each suited to different types of data and objectives. The process generally begins with a learning phase, where the model is constructed from available data. This is followed by an evaluation phase, in which the model's accuracy and reliability are rigorously tested using various performance metrics in a validation or testing phase. Upon successful validation, the model is deployed to make predictions on new, unseen data.

A well--designed predictive model must balance key factors such as performance, complexity, and interpretability. These considerations are essential to ensure that the model is not only effective but also reliable and suitable for practical application in real-world scenarios \citep{Barragan-Montero2022}. 

Performance is critical to guarantee that the predictive model is reliable. Thus, many measures to evaluate performance have been proposed over the years. In the context of diagnosis or classification systems, where the prediction is a nominal value, performance metrics often focus on assessing the accuracy of these categorical predictions. Commonly used metrics include accuracy, precision, recall, and F$_1$-score, each providing unique insights into different aspects of model effectiveness.

For instance, accuracy measures the overall correctness of predictions, while precision focuses on the proportion of correctly predicted positive cases out of all predicted positive cases. Recall, on the other hand, measures the model’s ability to identify all relevant positive cases, and the F$_1$-score balances precision and recall, offering a more holistic view of performance. Selecting the appropriate metric depends on the specific application and the importance of minimizing certain types of errors, such as false positives or false negatives, which can be critical in fields like healthcare.

In addition to these popular metrics, a measure that has gained relevance in recent years due to its properties is the Matthews Correlation Coefficient (henceforth, $\mcc$) \citep{Matthews1975}. Unlike other metrics, such as accuracy---which can be misleading when one class is significantly more prevalent than the other---the $\mcc$ evaluates true positives, true negatives, false positives, and false negatives in a balanced manner, providing a more comprehensive representation of the model's predictive capability in binary classification tasks. Nevertheless, its use has also sparked debate and some controversy, with researchers questioning its interpretation and applicability in certain contexts.

Now, as Matthews' seminal article reaches its 50th anniversary, it is an ideal moment to conduct a retrospective analysis of this remarkable measure, tracing its journey from its origins. 

The aim of this work is threefold: first, to conduct an objective examination of the historical role of the $\mcc$ in scientific research; second, to provide a rigorous exploration of its mathematical definition and properties, clarifying misconceptions as well as highlighting strengths and limitations which may help researchers in deciding whether to adopt or reject the $\mcc$ in favor of alternative metrics for classification problems; and third, to introduce analytical and graphical frameworks that not only deepen understanding of the $\mcc$ but also establish a general methodology for studying and comparing any classification performance measure.

The rest of the paper is organized as follows: Section \ref{sec:historical} offers a historical overview of the $\mcc$, distinguishing its origin, its uptake in structural biology, and its broader popularization. Section \ref{sec:mathematical} develops the mathematical overview: it traces the conceptual background of correlation, formalizes the $\mcc$, and revisits its equivalence to the Pearson coefficient and its relationship with the $\chi^2$ test; it then summarizes key properties and includes a dedicated subsection on statistical significance. Section \ref{sec:imbalance} turns to class imbalance, setting definitions and transformations, proving that $\mcc$ is not prevalence-invariant, and presenting a contour-based analysis that compares several measures across imbalance regimes to reveal bias, robustness, and sensitivity patterns. Finally, Section \ref{sec:conclusions} concludes and outlines directions for future work.

\section{Historical overview}\label{sec:historical}

The history of the $\mcc$ can be divided into three stages: its origin (1975), its application in structural biology (1975--1999), and its widespread popularization (2000--present).

\subsection{Origin and publication: 1975}

Brian W. Matthews is an emeritus biochemist and biophysicist from the University of Oregon, USA, renowned for his extensive and influential  publications in the field of structural biology, particularly in protein structure, function, and folding. His research has significantly advanced our understanding of protein behavior at the molecular level, making him a prominent figure in biochemistry and biophysics. However, among computer scientists, Matthews is primarily recognized for a single but impactful contribution: the introduction of the $\mcc$ in his paper entitled ``Comparison of the predicted and observed secondary structure of T4 phage lysozyme''.

Matthews' seminal article was submitted to Biochimica et Biophysica Acta on May 21, 1975, and was published later that year on October 20---nearly 50 years ago. This groundbreaking work introduced the $\mcc$, a measure that has since become a cornerstone in evaluating binary classification models and has subsequently been extended to multiclass contexts \citep{gorodkin2004}.

\subsection{Primary use in Structural Biology: 1975-1999}

During this period, the $\mcc$ found its primary application in structural biology, particularly in studies related to protein structure prediction. The $\mcc$ was mainly valuable in evaluating algorithms that predicted secondary structures of proteins, such as alpha-helices and beta-sheets. Throughout the late 1980s and 1990s, the $\mcc$ became a standard evaluation metric in structural biology for benchmarking predictive algorithms. The metric’s ability to provide a single, interpretable score, made it a preferred choice among researchers for comparing different predictive models.

According to Scopus, during 1975--1990 the article attracted only tens of citations, exclusively from structural biology (e.g., \citep{Argos1976,Nagano1977,Argos1982,Nishikawa1983}, among others). It was not until 1990 that a paper on the construction of cellular automata from experimental data referenced the correlation coefficient (without explicitly naming Matthews as the author) \citep{Richards1990}, plotting the correlation over time (cf. Fig. 12). Later, in 1992, a study on ultrasonic image segmentation using neural networks employed the correlation coefficient to evaluate method performance \citep{Prater1992}, even including a graph with a $y$-axis labeled as the ``Mat[t]hews coefficient'' (cf. Fig. 27). These instances marked the beginning of the $\mcc$ being used beyond structural biology, extending its application to other fields, and published in Physica D: Nonlinear Phenomena, and Ultrasonic Imaging journals, respectively. 

The first appearance of the $\mcc$ in a purely Computer Science journal was in Machine Learning in 1993, in an article by Cost and Salzberg \citep{Cost1993} entitled ``A Weighted Nearest Neighbor Algorithm for Learning with Symbolic Features''. This paper presents a nearest neighbor algorithm designed for contexts with categorical variables, using the Value Difference Metric, proposed by Stanfill and Waltz \citep{Stanfill1986}. The study addresses three distinct problems: predicting protein secondary structure, identifying DNA promoter sequences, and pronouncing English text. The $\mcc$ is only applied in evaluating the first problem, protein secondary structure prediction---still within structural biology.

Matthews’ article continued to be widely cited in structural biology (e.g., \citep{Brunak1991,Rost1993,Parodi1994,Nielsen1997,Blom1999}). Its visibility increased further with the emergence, in the late 1980s and early 1990s, of neural-network methods for protein--structure prediction and protein--pattern classification (e.g., \citep{Qian1988,Kneller1990,Stolorz1992,Metfessel1993}). Although neural networks were still in the early stages of adoption, researchers in these areas employed the $\mcc$ to evaluate their models, reflecting the complexity typical of biological datasets.

In the context of structural biology, some researchers analyzed the $\mcc$ and proposed new measures, such as the \textit{approximate correlation} (AC) \citep{Burset1996}, based on the \textit{average conditional probability} (ACP) \citep{Anderberg1973}, in order to measure the accuracy of predictions for protein structure. However, the $\mcc$ was already establishing its advantage over other evaluation measures in this discipline.

\subsection{Popularization: 2000--present}

The $\mcc$ began to gain widespread popularity from the early 2000s, largely due to the influential paper by Baldi et al., titled ``Assessing the accuracy of prediction algorithms for classification: An overview'' \citep{Baldi2000}. This landmark work examined the strengths and limitations of various evaluation metrics for predictive algorithms, underscoring the utility and robustness of the $\mcc$. The attention it brought to the coefficient’s advantages helped drive its broader adoption across diverse scientific fields.

Between 2000 and 2010, the use of the $\mcc$ broadened and consolidated across several fronts. In protein sequence and structure prediction, it became a routine evaluation criterion for multiple tasks (\citep{Hua2001,Garian2001,Yuan2002,Dyrlow2004,Chu2006,Kaur2004,Shen2007,Ghosh2008,Gao2007}). Driven by the expansion of omics, the $\mcc$ emerged as a standard metric across diverse classification problems (\citep{Steiner2004,Willenbrock2004,Liu2009,Yousef2008,Kaczkowski2009,Sakiyama2008}). In methodological studies, the $\mcc$ was likewise employed to benchmark and compare classification approaches (\citep{Guermeur2002,Guermeur2004,Green2006,Tsakonas2004,Jiang2006,Murakami2010,Worachartcheewan2010}).

The popularity of the $\mcc$ also spread to other disciplines, including finance \citep{Cipollini2009,Ghoddusi2019} and spatial data analysis \citep{Bosse2009,Wang2010,Amin2025}; and, with increasing frequency, to medicine \citep{Xu2015,Lu2018,ThorsenMeyer2020,Chan2025,Zhou2025,Cai2026} and machine learning \citep{Davison2018,Yao2020,Li2022}, ultimately becoming established in the latter and incorporated into analytical comparative studies of classification measures \citep{Jurman2012,Jiao2016,Tharwat2018,Chicco2020,Hicks2022}.

\section{Mathematical foundations}\label{sec:mathematical}

This section takes the introduction of the $\mcc$ as a starting point, but it is necessary to undertake a retrospective exploration to investigate the mathematical origins of correlation and correlation coefficients. With this approach, Subsection \ref{subsec:retrospective} addresses an in--depth look at the origin of the $\mcc$, giving credit to other researchers who made important earlier contributions. Among them, Karl Pearson was a pioneer in the study of correlation, so in Subsections \ref{subsec:intro_mcc}  and \ref{subsec:standard_formulation} the mathematical foundations of the $\mcc$ will first be presented, followed in Subsection \ref{subsec:pearson} reviews Pearson's coefficient \citep{Pearson1896} and its equivalence to the $\mcc$. In addition, there is a strong relationship between the $\mcc$ and the $\chi^2$ test \citep{Helmert1876}, which is described in Subsection \ref{subsec:chi2}. Finally, interesting properties of the $\mcc$ will be discussed in Subsection \ref{subsec:properties}, followed by a brief analysis of its statistical significance in Subsection \ref{subsec:statistics}.

\subsection{Background on the concept of correlation}\label{subsec:retrospective}

The term \textit{correlation} was first introduced in a paper read to the Royal Society of London on December 5, 1888, by Francis Galton, entitled ``Co-relations and their measurement, chiefly from anthropometric data'' \citep{Galton1888}. Galton did not, however, recognize the possibility of negative correlation values, nor their interpretation. This concept was introduced shortly thereafter by Raphael Weldon in 1890, who explicitly discussed the meaning of \textit{negative correlation} in the Proceedings of the Royal Society (Vol. XLVII, p. 445). Nevertheless, the term correlation, with extrinsic meaning, was already used earlier in French (e.g., in 1844 \citep{Bravais1844}; cf. ``corrélation'' on page 9).

According to Rodgers and Nicewander \citep{Rodgers1988}, the mathematical foundations for dealing with correlated variables were already present in the work of Carl Friedrich Gauss in 1823 \citep{Gauss1823}, who developed the so-called normal surface for $m$ \textit{correlated} variables. However, Gauss did not focus on correlation as an independent concept, but rather as a parameter embedded in his broader distributional framework.

In 1909, Franz Boas published a paper presenting a formula for the coefficient of correlation~\citep{Boas1909}. A few months later, Karl Pearson responded~\citep{Pearson1909a}, arguing that Boas' formulation had already been proposed by himself nearly a decade earlier~\citep{Pearson1900a}. In fact, the mathematical expression Pearson published in 1900 (cf. Eq. XXXVIII, p. 12) for the correlation coefficient in a fourfold table (see Table~\ref{tab:Pearson_table} and Eq.~\ref{eq:Pearson_1900}) is structurally similar to the formulation proposed a century later by Baldi et al.~\citep{Baldi2000} (Eq. \ref{eq:Baldi_2000}).
\begin{table}[tp]
\begin{center}
\caption{\mbox{Pearson's contingency table.}}
\label{tab:Pearson_table}
\begin{tabular}{cc|c}
\toprule
 $a$  &  $b$ & $a+b$ \\ 
$c$  & $d$ & $c+d$ \\ 
 \midrule
$a+c$ & $b+d$ & $n$ \\ 
\bottomrule 
\end{tabular}
\end{center}
\end{table}
\begin{equation}\label{eq:Pearson_1900}
r=\frac{ad-bc}{\sqrt{(b+d)(a+c)(c+d)(a+b)}}
\end{equation}

The concept of correlation and its associated coefficients has multiple, intertwined historical origins. Yule~\citep{Yule1896,Yule1909}, who had already emphasized the significance of Bravais’ foundational work on correlation~\citep{Bravais1844}, went on to propose his own measure---the coefficient of association Q---in 1900~\citep{Yule1900} for binary data. Consequently, the Pearson correlation coefficient is sometimes referred to as the Bravais--Pearson coefficient; notably, Karl Pearson introduced his version of the correlation coefficient in 1896~\citep{Pearson1896}, explicitly citing Bravais. For $2\times2$ contingency tables, the \emph{phi coefficient} ($\phi$) emerged as the binary analogue of Pearson correlation, as it is exactly the Pearson correlation between two dichotomous variables (coded 0/1)~\citep{Yule1900}. 

While Britain saw intense research activity on Statistics during 1865--1930 \citep{Mackenzie1981}, in 1887 the Bulletin of the Philosophical Society of Washington (Vol. X), published in the Bulletin of the Mathematical Section (24th meeting, February 16) the lecture by M. H. Doolittle, named ``Association ratios'', which had been already published in 1884 with the title ``The verification of predictions''  \citep{Doolittle1884}. In this work, Doolittle described the elements of a contingency table (without depicting it), as in Table \ref{tab:Doolittle}, from which he derived the \textit{discriminate association ratio---termed ``value of vitiated probability of the skillful predictions''---} illustrated in Eq. \ref{eq:Doolittle} (see an interesting discussion in \citep{Armistead2016}). This expression is exactly the square of the $\mcc$, and it does not consider the sign of the coefficient, but only the magnitude of the correlation.
\begin{table}[tp]
\begin{center}
\caption{\mbox{Doolittle's contingency table.}}
\label{tab:Doolittle}
\begin{tabular}{c|cc|c}
 & $A$  &  $\overline{A}$ &  \\ 
\toprule
$B$ & $c$  &  $b-c$ & $b$ \\ 
$\overline{B}$ & $a-c$  & $n-a-b+c$ & $n-b$ \\ 
 \midrule
& $a$ & $n-a$ & $n$ \\ 
\bottomrule 
\end{tabular}
\end{center}
\end{table}
\begin{equation}\label{eq:Doolittle}
\textrm{Doolittle's ratio}=\frac{(cn-ab)^2}{ab(n-a)(n-b)}
\end{equation}

The reader can find insightful discussions on the concept, evolution, and interpretations of correlation in several historical and methodological studies~\citep{Pierce1884,Yule1897,Yule1912,Pearson1913,Hubalek1982,Rodgers1988,Warrens2008,Beh2024}.

\subsection{Definition of the $\mcc$}\label{subsec:intro_mcc}

Matthews introduced the correlation coefficient within the context of prediction of the secondary structure of T4 phage lysozyme on the basis of the amino acid sequence, considering protein secondary structures, and in particular, determining for each residue whether or not it is in a secondary structure (e.g., a helix). A comparison between prediction and observed structure defines four categories: $p$ (predicted helical, observed helical), $q$ (predicted non-helical, observed non-helical), $u$ (predicted non-helical, observed helical; i.e., underpredicted) and $v$ (predicted helical, observed non-helical; i.e., overpredicted). These categories can be represented in a $2\times 2$ contingency matrix (Table \ref{tab:contingency2x2}).
\begin{table}[tp]
\begin{center}
\caption{\mbox{Predicted ($P$) in columns; Observed ($S$) in rows.}}
\label{tab:contingency2x2}
\begin{tabular}{c|cc}
\toprule
 & 1 & 0 \\
\midrule
1 & $p$ & $u$ \\ 
0 & $v$ & $q$ \\
\bottomrule 
\end{tabular}
\end{center}
\end{table}

Matthews observed that some measures of agreement between prediction and observation were not appropriate to this task \citep{Chou1974,Kabat1974}, and chose the correlation (described in Fisher's book \citep{Fisher1958}; cf. Ch. VI, p. 184), but expressed in terms of the prediction $P$ and observation $S$ as in Eq. \ref{eq:correlation}.
\begin{equation}\label{eq:correlation}
C = \frac{\sum_{i=1}^{n} (S_i - \bar{S})(P_i - \bar{P})}{\sqrt{\left(\sum_{i=1}^{n} (S_i - \bar{S})^2\right) \left(\sum_{i=1}^{n} (P_i - \bar{P})^2\right)}}
\end{equation}
In the special case where $P$ and $S$ are step functions with values of unity or zero, Eq. \ref{eq:correlation} can be transformed into Eq. \ref{eq:Matthews}.
\begin{equation}\label{eq:Matthews}
C = \frac{p/n - \bar{P}\bar{S}}{\sqrt{\bar{P}\bar{S}(1-\bar{S})(1-\bar{P})}} 
\end{equation}
where $\bar{S}$ is the fraction of positive observations, $\bar{P}$ is the fraction of positive predictions, and $n$ the total amount of observations $n=p+q+u+v$. 
\begin{equation*}\label{eq:SP}
\bar{S} = \frac{p+u}{n} \quad \bar{P} = \frac{p+v}{n}  
\end{equation*}

\subsection{Standard mathematical formulation}\label{subsec:standard_formulation}

\begin{definition}[Indicator function]\label{def:indicator}
Let \( X \) be a set, and let \( A \subseteq X \) be a subset of \( X \). The \textit{indicator function} (also known as the characteristic function) of \( A \) is the function $\mathbb{I}_A : X \to \{0,1\}$ defined by:
\begin{equation}
\mathbb{I}_A(x) =
\begin{cases}
1 & \text{if } x \in A \\
0 & \text{otherwise}
\end{cases}
\qquad
\mathbb{I}[C] =
\begin{cases}
1 & \text{if } C \text{ is true} \\
0 & \text{if } C \text{ is false}
\end{cases}
\end{equation}
In practice, especially in statistics and machine learning, it is common to use the notation \( \mathbb{I}[C] \) to denote the indicator function applied to a logical condition \( C \), in such a way that it generalizes naturally to logical predicates, where \( \mathbb{I}[C] \) can be interpreted as \( \mathbb{I}_{\{x \in X : C(x)\}} \).
\end{definition}
In the binary classification context, let $\mathrm{Y} = 1$ for the positive class and $\mathrm{Y} = 0$ for the negative class in the actual labels; similarly, $\hat{\mathrm{Y}} = 1$ for the predicted positive class and $\hat{\mathrm{Y}} = 0$ for the predicted negative class.
\begin{definition}[TP, FN, FP, TN]
Let $Y, \hat{Y} \in \{0,1\}^n$ be the true and predicted label vectors of size $n\in\mathbb{N}$, the values of TP, TN, FP and FN can be defined as follows:
\[
\begin{aligned}
TP(Y,\hat{Y}) &= \sum_{i=1}^n \mathbb{I}[y_i = 1 \land \hat{y}_i = 1] \qquad \qquad
FN(Y,\hat{Y}) = \sum_{i=1}^n \mathbb{I}[y_i = 1 \land \hat{y}_i = 0] \\
FP(Y,\hat{Y}) &= \sum_{i=1}^n \mathbb{I}[y_i = 0 \land \hat{y}_i = 1] \qquad \qquad
TN(Y,\hat{Y}) = \sum_{i=1}^n \mathbb{I}[y_i = 0 \land \hat{y}_i = 0]
\end{aligned}
\]
\noindent where $TP$, $FP$, $FN$, and $TN$, stand for True Positives, False Positives, False Negatives, and True Negatives, respectively.

Here $ Y=(y_1,\dots,y_n)$ is the vector of true labels,  
$\hat{ Y}=(\hat{y}_1,\dots,\hat{y}_n)$ the predicted labels,  
and $\mathbb{I}[\cdot]$ is the indicator function introduced in Def. \ref{def:indicator}.  
Each function, defined in the space $ \{0,1\}^n \times  \{0,1\}^n \to \mathbb{N}_0$, returns a non--negative integer, counting how many sample indices satisfy the indicated logical condition.
\end{definition}
Usually, the values of TP, FN, FP and TN, are represented in the so-called \textit{confusion matrix}, as shown in Table \ref{tab:2x2_cm}. Although, the purpose and meaning of the confusion matrix is exactly the same as the contingency matrix---appeared much earlier---, the research community has finally adopted a term that was specifically coined for counting the female subject's perceptual \textit{confusions} among sixteen English consonants in presence of noise \citep{Miller1955}. Without reasonable justification beyond popularity, we will keep the term hereinafter.
\begin{table}[tp]
\begin{center}
\caption{\mbox{Reference for $2\times 2$ confusion matrix.}}
\label{tab:2x2_cm}
\begin{tabular}{c|cc}
\toprule
 & $\hat{\textrm{Y}}=1$ & $\hat{\textrm{Y}}=0$  \\
  \midrule
$\textrm{Y}=1$ & $\textrm{TP}$  & $\textrm{FN}$  \\
$\textrm{Y}=0$ & $\textrm{FP}$  & $\textrm{TN}$  \\  
\bottomrule 
\end{tabular}
\end{center}
\end{table}

In 2000, Baldi et al. \citep{Baldi2000} derived from the $2\times 2$ confusion matrix (similar to Table \ref{tab:2x2_cm}), the well-known expression used nowadays  shown in Eq. \ref{eq:Baldi_2000}, which is equivalent to Eq. \ref{eq:Pearson_1900}.
\begin{equation}\label{eq:Baldi_2000}
\mcc=\frac{\textrm{TP}\cdot \textrm{TN} - \textrm{FP}\cdot \textrm{FN}}{\sqrt{(\textrm{TP+FN})(\textrm{TP+FP})(\textrm{TN+FP})(\textrm{TN+FN})}}
\end{equation}

\subsection{Equivalence to Pearson's correlation coefficient}\label{subsec:pearson}

Ronald A. Fisher published the first edition of ``Statistical Methods for Research Workers'' in 1925, in which he explained that the correlation coefficient, or product moment correlation, could be derived from the consideration that the population correlation is the geometric mean of the two regression coefficients, $\tfrac{S(xy)}{S(x^2)}$ and $\tfrac{S(xy)}{S(y^2)}$, corresponding to the slopes of the two linear regressions (the slope of $Y$ on $X$ and of $X$ on $Y$, respectively). In this expression, $S(xy)$ denotes the covariance of $x$ and $y$, while $S(x^2)$ and $S(y^2)$ denote the variances of $x$ and $y$, respectively.

\begin{equation}\label{eq:Fisher}
r = \sqrt{\frac{S(xy)}{S(x^2)}\frac{S(xy)}{S(y^2)}}
\end{equation}

The ratio $\frac{S(xy)}{S(x^2)}$ represents the slope of the best-fit line in a simple linear regression of $\mathrm{Y}$ on $\mathrm{X}$:
\begin{equation}
\mathrm{Y} = \beta_0 + \beta_1 \mathrm{X} + \epsilon
\end{equation}
\noindent where $\beta_0$ is the intercept; $\beta_1 = \frac{\mathrm{Cov}(\mathrm{X}, \mathrm{Y})}{\mathrm{Var}(\mathrm{X})}$ is the slope, quantifying how much $\mathrm{Y}$ changes on average for a one-unit change in $\mathrm{X}$; and $\epsilon$ is the error term. Equivalently, the ratio $\frac{S(xy)}{S(y^2)}$ represents the slope of $\mathrm{X}$ on $\mathrm{Y}$. 

This slope represents how the probability of $\mathrm{Y} = 1$ changes as $\mathrm{X}$ changes, i.e., $\beta_1$ (regression of $\mathrm{Y}$ on $\mathrm{X}$) can be interpreted as the change in the likelihood of the positive outcome for $\mathrm{Y}$ given the value of $\mathrm{X}$. 

Henceforth, let us assume that the two binary variables will be: $\mathrm{Y}$ (actual values) and $\hat{\mathrm{Y}}$ (predicted values). Thus, the Pearson correlation coefficient $r$ between two binary variables $\mathrm{Y}$ and $\hat{\mathrm{Y}}$ can be expressed as:
\begin{equation}\label{eq:pearson_correlation}
r(\mathrm{Y}, \hat{\mathrm{Y}}) = \frac{\text{Cov}(\textrm{Y}, \hat{\textrm{Y}})}{\sqrt{\text{Var}(\mathrm{Y}) \cdot \text{Var}(\hat{\mathrm{Y}})}}
\end{equation}
where $\text{Var}(\mathrm{Y})$ and $\text{Var}(\hat{\mathrm{Y}})$ are the variances of the actual and predicted values, and $\text{Cov}(\textrm{Y}, \hat{\textrm{Y}})$ is the covariance of $\mathrm{Y}$ and $\hat{\mathrm{Y}}$.

Taking as reference Table \ref{tab:2x2_cm}, the total number of samples $\textrm{n}$ is then $\textrm{n = TP + TN + FP + FN}$. For binary variables, the mean (or expected value) of $\textrm{Y}$,  $\mathbb{E}[\mathrm{Y}]$, represents the proportion of positive cases in the actual labels, and similarly, the expected value of $\hat{\textrm{Y}}$, $\mathbb{E}[\hat{\mathrm{Y}}]$, represents the proportion of predicted values. Thus,
\begin{equation}
\mathbb{E}[\textrm{Y}] = \frac{\textrm{TP + FN}}{\textrm{n}}
\qquad
\mathbb{E}[\hat{\textrm{Y}}] = \frac{\textrm{TP + FP}}{\textrm{n}}
\end{equation}

The variance for a binary variable $\textrm{Y}$ is given by $\text{Var}(\textrm{Y}) = \mathbb{E}[\textrm{Y}](1 - \mathbb{E}[\textrm{Y}])$. Thus:
\begin{equation}
\text{Var}(\textrm{Y}) = \frac{(\textrm{TP + FN})}{\textrm{n}} \cdot \left(1 - \frac{\textrm{TP + FN}}{\textrm{n}}\right) = \frac{(\textrm{TP + FN})(\textrm{TN + FP})}{\textrm{n}^2}
\end{equation}

Similarly, for $\hat{\textrm{Y}}$:
\begin{equation}
\text{Var}(\hat{\textrm{Y}}) = \frac{(\textrm{TP + FP})(\textrm{TN + FN})}{\textrm{n}^2}
\end{equation}

The covariance $\text{Cov}(\textrm{Y}, \hat{\textrm{Y}})$ is given by:
\begin{equation}
\text{Cov}(\textrm{Y}, \hat{\textrm{Y}}) = \mathbb{E}[(\mathrm{Y} - \mathbb{E}[\mathrm{Y}])(\hat{\mathrm{Y}} - \mathbb{E}[\hat{\mathrm{Y}}])] = \mathbb{E}[\textrm{Y} \cdot \hat{\textrm{Y}}] - \mathbb{E}[\textrm{Y}] \cdot \mathbb{E}[\hat{\textrm{Y}}]
\end{equation}

The term $\mathbb{E}[\textrm{Y} \cdot \hat{\textrm{Y}}]$ represents the probability that both $\textrm{Y}$ and $\hat{\textrm{Y}}$ are 1:
\begin{equation}
\mathbb{E}[\textrm{Y} \cdot \hat{\textrm{Y}}] = \frac{\textrm{TP}}{\textrm{n}}
\end{equation}

Substituting:
\begin{equation}
\text{Cov}(\textrm{Y}, \hat{\textrm{Y}}) = \frac{\textrm{TP}}{\textrm{n}} - \frac{(\textrm{TP + FN})(\textrm{TP + FP})}{\textrm{n}^2}
\end{equation}

Now we can substitute these values into the Pearson correlation formula depicted in Eq. \ref{eq:pearson_correlation}. Plugging in the covariance and variance terms derived above:
\begin{equation}
r(\textrm{Y}, \hat{\textrm{Y}}) = \frac{\frac{\textrm{TP}}{\textrm{n}} - \frac{(\textrm{TP + FN})(\textrm{TP + FP})}{\textrm{n}^2}}{\sqrt{\frac{(\textrm{TP + FN})(\textrm{TN + FP})}{\textrm{n}^2} \cdot \frac{(\textrm{TP + FP})(\textrm{TN + FN})}{\textrm{n}^2}}}
\end{equation}

Multiplying through by $\textrm{n}^2$ in both the numerator and the denominator:
\begin{equation}
r(\textrm{Y}, \hat{\textrm{Y}}) = \frac{\textrm{n} \cdot \textrm{TP} - (\textrm{TP + FN})(\textrm{TP + FP})}{\sqrt{(\textrm{TP + FN})(\textrm{TN + FP})(\textrm{TP + FP})(\textrm{TN + FN})}}
\end{equation}

Finally, substituting $\textrm{n}$, it is exactly the formula for the Matthews Correlation Coefficient seen above in Eq. \ref{eq:Baldi_2000}:
\begin{equation}
 r(\textrm{Y}, \hat{\textrm{Y}}) = \frac{\textrm{TP} \cdot \textrm{TN} - \textrm{FP} \cdot \textrm{FN}}{\sqrt{(\textrm{TP + FP})(\textrm{TP + FN})(\textrm{TN + FP})(\textrm{TN + FN})}} = \mcc 
\end{equation}

The derivation above shows that the $\mcc$ is mathematically equivalent to the Pearson correlation coefficient calculated between binary actual and predicted values. This makes the $\mcc$ a specific case of Pearson correlation tailored to binary classification, accounting for true positives, true negatives, false positives, and false negatives.

\subsection{Relationship to $\chi^2$}\label{subsec:chi2}

While the chi-square ($\chi^2$) test \citep{Helmert1876} and the $\mcc$ serve different purposes, they can be related under specific contexts, particularly in binary classification. The $\chi^2$ test is used to assess whether there is a significant association between two categorical variables \citep{Pearson1900a,Yule1900, Mirkin2001}. For a $2 \times 2$ contingency table used in binary classification (Table \ref{tab:2x2_cm}), the test statistic is:
\begin{equation}
\chi^2 = \sum_{i=1}^2 \sum_{j=1}^2 \frac{(O_{ij} - E_{ij})^2}{E_{ij}}
\end{equation}
where $O_{ij}$ is the observed frequency for cell $(i, j)$, and $E_{ij}$ is the expected frequency under the null hypothesis of independence.

The observed frequencies $O_{ij}$ correspond directly to the entries of the confusion matrix:
\begin{align*}
O_{11} &= TP & O_{12} &= FN \\
O_{21} &= FP & O_{22} &= TN
\end{align*}

The expected frequencies $E_{ij}$, under the null hypothesis of independence between predicted and actual classes, are calculated using row and column totals:
\begin{align*}
R_1 &= TP + FN & R_2 &= FP + TN \\ 
C_1 &= TP + FP & C_2 &= FN + TN
\end{align*}

Then:
\begin{align*}
E_{11} &= \frac{R_1 C_1}{n} = \frac{(TP + FN)(TP + FP)}{n} \qquad
E_{12} = \frac{R_1 C_2}{n} = \frac{(TP + FN)(FN + TN)}{n} \\
E_{21} &= \frac{R_2 C_1}{n} = \frac{(FP + TN)(TP + FP)}{n} \qquad
E_{22} = \frac{R_2 C_2}{n} = \frac{(FP + TN)(FN + TN)}{n}
\end{align*}
    
Substituting the entries from observed and expected frequencies, expressed in terms of the confusion matrix, yields the closed form of $\chi^2$: 
\begin{equation}
\chi^2 = \frac{n(TP \cdot TN - FP \cdot FN)^2}{(TP + FP)(TP + FN)(TN + FP)(TN + FN)} = n\cdot \mcc^2
\end{equation}

Therefore, the $\chi^2$ test and $\mcc$ are related through the Pearson correlation coefficient in binary classification. $\chi^2$ measures how far the observed frequencies deviate from the expected frequencies under the assumption of independence, and $\mcc$ measures the strength and direction of the linear relationship between predicted and actual labels in binary classification. This connection highlights the statistical foundation of $\mcc$ and provides a pathway to assess its significance using the $\chi^2$ distribution.

Extensions of the correlation coefficient for larger contingency tables were later developed, including Tschuprow’s T~\citep{Tschuprow1939} and Cramér's V~\citep{Cramer1946}, both designed to normalize the $\chi^2$ statistic to account for table size. 

\subsection{Core properties} \label{subsec:properties}

In mathematical terms, the concepts of \textit{measure} and \textit{metric} are rigorously defined and fundamentally different. A measure is a function defined over a $\sigma$-algebra of subsets of a given set, satisfying properties such as non-negativity, null empty set, and countable additivity, and is typically used to quantify a property of sets. Conversely, a metric is a function defined over pairs of elements in a set that quantifies the \textit{distance} between them, adhering to axioms such as non-negativity, symmetry, identity of indiscernibles, and the triangle inequality. Despite these formal differences, in the field of machine learning, both terms are often used interchangeably to refer to any function that quantifies the discrepancy between a model's predictions and true labels. This pragmatic use reflects the focus on the practical utility of such functions as evaluation tools, rather than their strict mathematical classification. In this paper, while acknowledging the theoretical distinction, we will adopt this common practice and use the terms measure and metric with the same goal and meaning, referring generically to functions that assess classifier performance. Other terms like \textit{association} or \textit{agreement} coefficients are also employed \citep{Krippendorff1987}.

We will only focus on contingency-based performance measures, i.e., on measures that are strictly based on the values of $\mathrm{TP,FN,FP,TN}$, unlike other probability-based measures which need additional information on the probabilities of assigning an instance to a class, such as the Brier score \citep{Brier1950}, the area under the ROC curve \citep{Peterson1954,Swets1973,Hanley1982,He2008b}, or the area under the MCP (Multiclass Classification Performance) curve \citep{Aguilar2022a,Aguilar2024a,Aguilar2024b,Aguilar2024c}.

In the following, some definitions that will help formalize certain mathematical aspects of the classification performance measures and, in particular, of the $\mcc$, will be included.

\begin{definition}[Dissimilarity]\label{def:dissimilarity}
Let \( X \) be a non-empty set. A function $\delta : X \times X \to [0, \infty)$ is called a \textit{dissimilarity function} if it satisfies:
\begin{enumerate}
    \item Non-negativity: \( \delta(x, y) \geq 0 \quad \forall x, y \in X \)
    \item Reflexivity: \( \delta(x, x) = 0 \quad \forall x \in X \)
\end{enumerate}
\end{definition}
Dissimilarity functions quantify how different two objects are, without necessarily forming a metric space. A function is called a \textit{distance} if it satisfies the same properties as a dissimilarity, and possibly others (e.g., symmetry), but may not meet all the criteria required to be a metric (see Def. \ref{def:metric}). Distances are often used in machine learning even when they lack full metric structure.

It is worth noting that to measure the quality of classification, the distance between the true labels and the predicted labels must be calculated. In case the classification is perfect (i.e., all predicted labels are equal to true labels), the results should be 0 (reflexivity in Def. \ref{def:dissimilarity}). Therefore, measures that behave as losses (0 best, 1 worst) are most consistent with reflexivity \citep{Hand97}. Many classification measures do not hold this property (e.g., F-score, accuracy, and the $\mcc$).

\begin{definition}[Metric]\label{def:metric}
A function $d : X \times X \to [0, \infty)$ is called a \textit{metric} on \( X \) if for all \( x, y, z \in X \), the following hold:
\begin{enumerate}
    \item Non-negativity: \( d(x, y) \geq 0 \)
    \item Identity of indiscernibles: \( d(x, y) = 0 \iff x = y \)
    \item Symmetry: \( d(x, y) = d(y, x) \)
    \item Triangle inequality: \( d(x, z) \leq d(x, y) + d(y, z) \)
\end{enumerate}
The pair \( (X, d) \) is then called a \textit{metric space}.
\end{definition}

\begin{definition}[Performance measure]Let $f : \{0,1\}^n \times \{0,1\}^n \rightarrow [0,\infty)$ be a general classification performance measure, where the first argument corresponds to the vector of $n$ true labels $\mathbf{Y}$ and the second to the $n$ predicted labels $\hat{\mathbf{Y}}$.
\end{definition}
A performance measure should be a dissimilarity function, but not necessarily a metric. For instance, the $\mcc$ is not even a dissimilarity function because it can take negative values ($\mcc \in [-1,1]$). When transformed into $\mcc^* = \frac{1-\mcc}{2}$, it is scaled into $[0,1]$ and can be interpreted as a dissimilarity function. However, even after transformation, it cannot be considered a metric, as it fails to satisfy the triangle inequality, although it remains symmetric.

\begin{definition}[Confusion matrix--based performance measure]A confusion matrix-based performance measure $\mathcal{M}$ is a real--valued function 
\begin{equation}\label{eq:measure}
\mathcal{M}:\;\bigl\{(\mathrm{TP},\mathrm{FN},\mathrm{FP},\mathrm{TN})\in\mathbb{N}_{0}^4\bigr\}\;\to\;\mathbb{R}
\end{equation}
such that $f(Y,\hat{Y})=\mathcal{M}(TP(Y,\hat{Y}),FN(Y,\hat{Y}),FP(Y,\hat{Y}),TN(Y,\hat{Y}))$, subject to the constraints that $\mathrm{TP,FN,FP,TN}$ are non-negative integer counts. 
\end{definition}

\begin{definition}[Completeness]A performance measure  $\mathcal{M}$ is complete iff its mathematical definition contains all the four elements of the domain  $(\mathrm{TP},\mathrm{FN},\mathrm{FP},\mathrm{TN})$.
\end{definition}
Unlike $\mathrm{F}_1$-score or Sensitivity, $\mcc$ uses all four cells of the confusion matrix explicitly.

\begin{definition}[Symmetry]A performance measure  $\mathcal{M}$ is symmetric iff exchanging the values of $(\mathrm{TP}, \mathrm{FN})$ with $(\mathrm{TN}, \mathrm{FP})$ yields the same magnitude.
\end{definition}

A key property of $\mcc$ is that it is symmetric with respect to positives versus negatives, unlike Precision and $F_1$-score.

\begin{definition}[Range]
Let $f:\;\{0,1\}^{n}\times\{0,1\}^{n}\;\longrightarrow\;[0,\infty)$ be a classification performance measure. The \textit{range} of $f$, denoted $\operatorname{Range}(f)$, is the set of all real values attained by~$f$ when evaluated on every admissible pair
of true- and predicted-label vectors:
\[
\operatorname{Range}(f)
\;=\;
\bigl\{\,f( Y,\hat{Y})
      :  Y,\hat{ Y}\in\{0,1\}^{n}\bigr\}.
\]
\end{definition}

For any $n\ge 2$ the $\mcc$,  
\[
\mcc:
\{0,1\}^{n}\times\{0,1\}^{n}\longrightarrow[-1,1]
\]
i.e., $\operatorname{Range}(\mcc)=[-1,1]$.

In particular, $\mcc\!=\!+1$ occurs when $\hat{Y}\!=\!Y$ (perfect prediction), and then $\mathrm{FP}=\mathrm{FN}=0$ and $\mathrm{TP},\mathrm{TN}>0$. $\mcc\!=\!-1$ occurs when $\hat{Y}=1- Y$ (complete class inversion) and the denominator is non-zero, and then $\mathrm{TP}=\mathrm{TN}=0$ and $\mathrm{FP}, \mathrm{FN}>0$. $\mcc\!=\!0$ if there is no correlation between actual and predicted labels. For large samples with near-random predictions, we often see $\mcc \approx 0$. This happens when the proportion of correct predictions on one class equals the proportion of incorrect predictions on the other. So even if TP and TN are not zero, they are balanced out by symmetric errors, i.e., when $\frac{TP}{FN}\approx \frac{FP}{TN}$.

It is well known that $\mcc\!=\!0$ is often described---rather loosely---as ``no better than random'', or even as ``predictions made by chance'' \citep{Chicco2023}. Formally, $\mcc\!=\!0$ occurs \emph{iff} the numerator of the coefficient vanishes, i.e.\ $TP\!\cdot\!TN = FP\!\cdot\!FN$. 

Let the True Positive Rate (TPR), the False Positive Rate (FPR), and the True Negative Rate (TNR) be defined as follows:
\begin{equation}\label{eq:tpr_fpr}
TPR = \frac{TP}{TP+FN}
\qquad
FPR = \frac{FP}{FP+TN}
\qquad
TNR = 1-FPR=\frac{TN}{FP+TN}
\end{equation}
For any non-degenerate dataset $\mcc\!=\!0$ is equivalent to the equality of rates $TPR\!=\!FPR$; the classifier (FPR--TPR) point therefore lies on the \emph{no-discrimination} diagonal. Crucially, $TPR = FPR$ does \emph{not} guarantee that the predictions themselves are stochastic: a fully deterministic model can still satisfy $TPR = FPR$ by favoring one class with identical error rates on both true classes. The sequence of balanced confusion matrices $(TP,FN,FP,TN)=\{(99,1,99,1),(98,2,98,2),\dots,(1,99,1,99)\}$ illustrates the point---each case yields $\mcc=0$ despite a strong, non-random bias toward one class. Likewise, dataset imbalance does not alter the fact that any point with $TPR = FPR$ produces $\mcc=0$; imbalance merely shifts other contour levels but leaves the no-discrimination line unchanged (see Fig. \ref{fig:metrics_contours} in Section \ref{sec:imbalance} devoted to imbalance). Hence $\mcc=0$ should be interpreted as ``no linear correlation between prediction and truth'', not necessarily as evidence that the classifier is random or makes predictions by chance.
		
On the other hand, the sign of $\mcc$ is determined by the relationship between $TPR$ and $FPR$, i.e., if $TPR\!>\!FPR$ then $\mcc\!>\!0$, and if $TPR\!<\!FPR$ then $\mcc\!<\!0$. 

\begin{definition}[Monotonicity] \label{def:monotone}
Let $\mathcal{M} : \mathbb{N}_{0}^{4}\!\longrightarrow\!\mathbb{R},\;
\mathcal{M}=\mathcal{M}(TP,FN,FP,TN)$ be a scalar performance measure
that depends only on the four confusion-matrix counts.
We say that $\mathcal{M}$ is \emph{component-wise monotone} iff it
satisfies that $\mathcal{M}(TP,FN,FP,TN) < \mathcal{M}(TP+1,FN-1,FP,TN)$ and $\mathcal{M}(TP,FN,FP,TN) < \mathcal{M}(TP,FN,FP-1,TN+1)$.
\end{definition}

The $\mcc$ satisfies this monotonicity. However, F-score generally does not.

\subsection{Statistical significance}\label{subsec:statistics}

Assessing the statistical significance of the $\mcc$ is essential when using it as a performance measure in classification tasks. Point estimates of the $\mcc$ provide useful summaries, but they do not account for sampling variability. Without a measure of uncertainty \citep{Aguilar2024f}, it is difficult to judge whether observed differences between models are genuine or merely the result of random variation. For this reason, it is important to accompany the $\mcc$ with confidence intervals (CI), thereby quantifying the reliability of the reported values.

A widely used method to obtain CIs for correlation-based statistics is Fisher’s $z$-transformation. By applying the transformation $z = \tanh^{-1}(\hat{r})$, the sample correlation $\hat{r}$ is approximately normally distributed with mean $\tanh^{-1}(\rho)$ (where $\rho$ denotes the population correlation) and variance $1/(n-3)$, where $n$ is the sample size. This leads to a straightforward closed-form CI. However, this approximation relies on the assumption of bivariate normality for $(\textrm{Y}, \hat{\textrm{Y}})$---which does not hold for binary outcomes---, and its reliability can be limited, often resulting in CIs with poor coverage (typically too narrow).

An alternative approach is based on $\chi^2$ inversion, which derives the CI by inverting a likelihood ratio test. Specifically, the log-likelihood of a multinomial $2\times 2$ table is used to compute the likelihood ratio statistic for candidate values of $\rho$. The CI consists of the set of correlations for which the likelihood ratio does not exceed the critical value of the $\chi^2_1$ distribution (1 degree of freedom). This method is computationally more intensive, requiring numerical optimization, but it provides intervals with more accurate coverage properties, especially in finite samples, and is thus generally considered a more rigorous alternative.

\begin{example}Consider a sample size of $n = 1000$ with an observed $\mcc$ of $\hat{r} = 0.65$. Using Fisher’s method, the 95\% CI is obtained by transforming $\hat{r}$ into the $z$-scale, computing $z = \tanh^{-1}(0.65) \approx 0.775$ with a standard error of $1/\sqrt{997} \approx 0.0317$. The interval in the $z$-scale is $[0.713, 0.837]$, which corresponds to $[0.613, 0.684]$ when back-transformed. By contrast, applying $\chi^2$ inversion yields a wider 95\% CI, approximately $[0.588, 0.712]$. This illustrates how Fisher's method may underestimate \emph{uncertainty}, whereas $\chi^2$ inversion provides a more conservative and reliable assessment of the statistical significance of the $\mcc$.
\end{example}

Finally, it is important to note that when the sample size is small or the class distribution is highly skewed (e.g., biomedical datasets), the reliability of asymptotic methods such as Fisher's $z$-transformation or $\chi^2$ inversion may be compromised. In such cases, a non-parametric bootstrap procedure provides a valuable alternative for assessing the stability of the $\mcc$ estimate. 

\section{Class imbalance}\label{sec:imbalance}

Class imbalance is the condition in which the class prior (or sample) distribution is strongly skewed---one or more minority classes occur substantially less often than others---in either binary or multiclass problems. In binary classification, this typically manifests as a rare minority class and a dominant majority class.

Class imbalance is a critical issue in machine learning evaluation because it biases both performance measures and the learning process itself \citep{Kubat1997}. In the context of binary classification, many widely used metrics can be highly misleading under skewed distributions, since they may conceal poor detection of the minority class behind the dominance of the majority class. More robust measures have therefore gained attention for their ability to reflect classifier performance across all outcomes of the confusion matrix. The implications of evaluating classifiers on imbalanced datasets are far-reaching, since misleading assessments may lead to the deployment of models that appear effective overall yet systematically fail in precisely those cases where reliable predictions are most needed (e.g., clinical diagnosis).

While numerous strategies have been proposed to mitigate imbalance---such as resampling \citep{Batista2004,He2009}, or synthetic data generation \citep{Chawla2002,Han2005,He2008a,Taccaliti2025}---the present work does not focus on these corrective techniques. Instead, our interest lies in the evaluation perspective, emphasizing how the choice of performance measure affects the interpretation of classifier behavior in imbalanced scenarios.

In binary datasets, class imbalance simply means $P\neq N$ (i.e., $\textrm{\emph{TP+FN}} \neq \textrm{\emph{FP+TN}}$); we usually speak of strong imbalance when the difference is substantial (e.g., $\textrm{\emph{TP+FN}} \gg \textrm{\emph{FP+TN}}$ or $\textrm{\emph{TP+FN}} \ll \textrm{\emph{FP+TN}}$). 

\begin{definition}[Imbalance ratio]
Let a binary data set contain $P$ positive and $N$ negative examples 
($P,N\in\mathbb{N}$).  
Define the \emph{imbalance ratio}
\[
\mathrm{IR}\;=\;\frac{\max\{P,N\}}{\min\{P,N\}}
    \;=\;\frac{|\,P-N\,|}{\min\{P,N\}}+1.
\]

A data set is \emph{perfectly balanced} iff \,$\mathrm{IR}=1$ (i.e.,\ $P=N$). The larger $\mathrm{IR}>1$, the more \emph{imbalanced} the class distribution.

\end{definition}

Class‑skewed ($\mathrm{IR}\gg1$) data can distort many performance measures (e.g., accuracy or F$_1$-score) because the majority class dominates the error counts. Metrics designed to be \emph{class-prevalence insensitive} should exhibit stability across different imbalance values.

An imbalance analysis consists in describing how the measure $\mathcal{M}$ changes if the number of actual positives ($\mathrm{TP}+\mathrm{FN}$) or actual negatives ($\mathrm{FP}+\mathrm{TN}$) are scaled, while preserving the classifier’s rate of correct classification on each class. If the dataset is rebalanced (e.g., up- or down-sampled) so that the ratio of actual positives to actual negatives changes, the interest resides in knowing whether $\mathcal{M}$ remains the same, provided the classifier's conditional accuracy on each class is unchanged.

\begin{definition}[Imbalance Transformation] It is a transformation 
$\mathrm{T}_{(\alpha,\beta)}$ for $\alpha,\beta\! \in\! \mathbb{N}$ as follows:
\[
\mathrm{T}_{(\alpha,\beta)}:\;
(\mathrm{TP},\mathrm{FN},\mathrm{FP},\mathrm{TN})
\;\;\mapsto\;\;
(\alpha\,\mathrm{TP},\;\alpha\,\mathrm{FN},\;\beta\,\mathrm{FP},\;\beta\,\mathrm{TN})
\]
so that all actual positive entries ($\mathrm{TP}$, $\mathrm{FN}$) are multiplied by $\alpha>0$ and all actual negative entries ($\mathrm{FP}$, $\mathrm{TN}$) are multiplied by $\beta>0$.
\end{definition}

\begin{definition}[Prevalence Invariance]
A measure $\mathcal{M}$ is said to be insensitive to imbalance (or prevalence--invariant) if:
\begin{equation}
\forall\; \alpha,\beta \in \mathbb{N},\quad
\mathcal{M}\Bigl(\mathrm{T}_{(\alpha,\beta)}(\mathrm{TP,FN,FP,TN})\Bigr)
\;=\;
\mathcal{M}(\mathrm{TP,FN,FP,TN}).
\end{equation}
Thus, if the row of actual positives and/or the row of actual negatives are scaled by arbitrary positive factors (thus changing how many examples belong to each class in the dataset), the measure $\mathcal{M}$ must remain the same---provided that $\mathrm{TPR}$ and $\mathrm{TNR}$ (Eq. \ref{eq:tpr_fpr}) do not change.
\end{definition}

\begin{theorem}[$\mcc$ is not insensitive to imbalance] \label{th:insensitive}
Applying $\mathrm{T}_{(\alpha,\beta)}$ to $\mcc$ becomes
\[
\mcc=\frac{(\alpha\,\mathrm{TP})(\beta\,\mathrm{TN}) - (\beta\,\mathrm{FP})(\alpha\,\mathrm{FN})}
{\sqrt{(\alpha\,\mathrm{TP}+\beta\,\mathrm{FP})(\alpha\,\mathrm{TP}+\alpha\,\mathrm{FN})(\beta\,\mathrm{TN}+\beta\,\mathrm{FP})(\beta\,\mathrm{TN}+\alpha\,\mathrm{FN})}},
\]
which simplifies to 
\[
\mcc=\frac{\bigl(\mathrm{TP\cdot TN - FP\cdot FN}\bigr)}
{\sqrt{ \left(\mathrm{TP}+\frac{\beta}{\alpha}\,\mathrm{FP}\right)\left(\mathrm{TP}+\mathrm{FN} \right)\left(\mathrm{TN}+\mathrm{FP} \right)\left(\mathrm{TN}+\frac{\alpha}{\beta}\mathrm{FN} \right)}}
\]
Thus, $\mcc= \mcc\Bigl(T_{(\alpha,\beta)}(\mathrm{TP, FN, FP, TN})\Bigr) \Leftrightarrow \frac{\beta}{\alpha}=\frac{\alpha}{\beta}=1 \Leftrightarrow \alpha=\beta$. Therefore,  $\mcc$ is not insensitive to imbalance. Equivalently, $\mcc$ is invariant only to uniform rescaling ($\alpha=\beta$).
\end{theorem}

\begin{example}
Assume actual positives = 3 ($\textrm{Y}=1$), actual negatives = 4 ($\textrm{Y}=0$), with the confusion matrix
\[
\begin{array}{c c c}
\begin{array}{c|cc}
 & \hat{\textrm{Y}}=1 & \hat{\textrm{Y}}=0\\
\hline
\textrm{Y}=1 & \mathrm{TP}=2 & \mathrm{FN}=1\\
\textrm{Y}=0 & \mathrm{FP}=1 & \mathrm{TN}=3
\end{array}
&
\displaystyle
\quad \mathrm{TPR}\! =\! \frac{\mathrm{TP}}{\mathrm{TP} + \mathrm{FN}} \!=\! \frac{2}{2+1}=\frac{2}{3}
&
\displaystyle
\quad \mathrm{TNR} \!= \!\frac{\mathrm{TN}}{\mathrm{TN} + \mathrm{FP}} \!= \!\frac{3}{3+1}=\frac{3}{4}
\end{array}
\]
\[
\mcc =\frac{(2\cdot 3) - (1\cdot 1)}{\sqrt{(2+1)(2+1)(3+1)(3+1)}}\approx 0.417
\]
Now multiplying the negative portion by 5, and leaving positives the same:
\[
\begin{array}{c c c}
\begin{array}{c|cc}
 & \hat{\textrm{Y}}=1 & \hat{\textrm{Y}}=0\\
\hline
\textrm{Y}=1 & \mathrm{TP}=2 & \mathrm{FN}=1\\
\textrm{Y}=0 & \mathrm{FP}=5 & \mathrm{TN}=15
\end{array}
&
\displaystyle
\quad\mathrm{TPR} \!=\! \frac{\mathrm{TP}}{\mathrm{TP} + \mathrm{FN}} \!=\! \frac{2}{2+1}\!=\!\frac{2}{3}
&
\displaystyle
\quad\mathrm{TNR} \!=\! \frac{\mathrm{TN}}{\mathrm{TN} + \mathrm{FP}} \!= \!\frac{15}{15+5}\!=\!\frac{3}{4}
\end{array}
\]
\[
  \mcc
  = \frac{(2\cdot 15) - (5\cdot 1)}{\sqrt{(2+5)(2+1)(15+5)(15+1)}}
  \approx 0.305
\]
Therefore, $\mcc$ dropped from 0.417 to 0.305, even though $\mathrm{TPR}$ and $\mathrm{TNR}$ stayed the same.
\end{example}

The very simple example above shows that $\mcc$ is sensitive to class imbalance. It is often said that $\mcc$ is robust, but only because it uses $\mathrm{(TP, FN, FP, TN)}$ in a symmetric fashion. Also, as the $\mcc$ is a correlation-based measure, it is less biased toward the majority class than simpler metrics like accuracy. However, for extreme imbalance, some terms like $(\mathrm{TP} + \mathrm{FP})$ or $(\mathrm{TN} + \mathrm{FN})$ can become very large or very small, making $\mcc$ values volatile or less straightforward to interpret.

$\mcc$ differs from measures such as Youden's $J$ \citep{Youden1950}, which depend only on the error rates ($\mathrm{TPR}$ and $\mathrm{TNR}$). Thus, $\mcc$ can shift if the distribution of predicted positives versus negatives changes at a different rate from the distribution of actual classes.

In general, the $\mcc$ can be problematic to interpret on highly imbalanced datasets, because it does not capture how well the classifier is performing for the positive class when it is performing wrong for the negative class. For instance, if $\mathrm{TN}=0$ and $\mathrm{FN}=0$, all the predictions are positive (true and false). In this case the $\mcc$ is undefined (often reported as 0 by software), while the accuracy still reports the proportion of good predictions, $\mathrm{Acc}=\frac{\mathrm{TP}}{\mathrm{TP}+\mathrm{FP}}$. 

\subsection{Impact of imbalance on $\mcc$}

In classification tasks, especially under class imbalance, traditional scalar summaries (such as reporting a single value of accuracy, F$_1$-score, or $\mcc$) may fail to capture how a metric responds to different types of classification errors. 

In general, the impact of imbalance on performance measures is remarkable, and even mild levels of imbalance---at 30--40\% of the data in the minority class---is sufficient to alter the values of the measures commonly used to assess performance \citep{Wei2013}.

Some recent, relevant publications still state that $\mcc$ is invariant to class imbalance \citep{Kaissis2021,Hanser2025,Jia2026}. Other authors encourage the scientific community to use the $\mcc$ instead of F$_1$-score \citep{Chicco2020}, the Cohen's $\kappa$ \citep{Chicco2021}, the Fowlkes--Mallows index \citep{Chicco2023}, the diagnostic odds ratio (DOR) \citep{Chicco2021b}, the balanced accuracy \citep{Chicco2021c}, and the area under the ROC curve \citep{chicco2023b}.  Although some authors state that the most robust measure for binary classification is $\mcc$ \citep{Canbek2022}, the robustness has only been examined via benchmarking methods \citep{Canbek2021}.

In contrast, a few authors state that the $\mcc$ is very sensitive to data composition \citep{Racz2019}, or prefer to omit the $\mcc$ as it is a discrete version of the Pearson's correlation coefficient \citep{Brusco2021,Silveira2023}, or because it is the geometric mean of other two measures: informedness and markedness \citep{Powers2011}. Foody \citep{Foody2023} argued that the $\mcc$ has been unrealistically promoted as an accuracy metric to summarize a binary classification and has some undesirable properties associated with prevalence and the use of an imperfect reference standard. Thus, high $\mcc$ score could potentially arise from a modest or even poor classification. For that reason, Zimmerman et al. \citep{Zimmerman2022} recommended to not endorse the use of the $\mcc$ as the metric for comparing testing methods. Although the $\mcc$ is usually a good performance metric in comparison to others (e.g., when 37 metrics are analyzed \citep{Yilmaz2023}), and experimental proofs of its suitability for imbalanced data have been shown when using a classifier specifically based on this metric \citep{Boughorbel2017}, many works raise important concerns because the effectiveness of $\mcc$ may be compromised in imbalanced datasets \citep{Foody2023,Itaya2025}. 

Zhu \citep{Zhu2020} analyzed the first--order partial derivatives of the $\mcc$ with respect to TN and TP and concluded that the $\mcc$ is affected by imbalance: the higher the imbalance ratio, the metric tends to be more skewed and exhibits non--linear behavior with respect to the linear increases of TP and TN values. This interesting fact has been analyzed in depth in this work, and it will be further discussed.

\subsection{Analytical study}

Contour plots provide a two-dimensional, continuous visualization that allows researchers to explore the behavior of performance metrics across the full range of possible classifier outcomes. This makes them particularly powerful for understanding the robustness, symmetry, and bias, and serve as a powerful diagnostic tool for uncovering how a given measure behaves under imbalance. Any explicit dependence of $\mathcal{M}$ on the absolute counts $P$ and $N$ (or their ratio $\mathrm{IR}$) will geometrically manifest in the contour plot as a distortion of the level curves compared to the balanced scenario ($\mathrm{IR}=1$). 

A contour plot maps the values of a classification metric over the $(\mathrm{FPR}, \mathrm{TPR})$ plane. Each point on this plane corresponds to a hypothetical classifier with a given FPR and TPR, obtained from the confusion matrix ($\TP$, $\FN$, $\FP$, $\TN$), and from which the metric of interest can be derived (given a fixed number of positive and negative instances). The contour lines (also called isocurves or level sets) connect all points in the plane that yield the same metric value. The resulting pattern reveals how a metric reacts to changes in $\FPR$ and $\TPR$---and, indirectly, to the trade-off between Type I and Type II errors.

For a binary data set with $P$ positives and $N$ negatives, a point $(\FPR,\TPR)\!\in\![0,1]^2$ induces
\begin{equation}
(\TP,\FN,\FP,\TN)
\!=\!\bigl(\TPR\cdot \mathrm{P},\;(1\!-\!\TPR)\cdot\mathrm{P},\;\FPR\cdot \mathrm{N},\;(1\!-\!\FPR)\cdot\mathrm{N}\bigr)
\end{equation}

If a performance measure $\mathcal{M}:\mathbb{N}_0^4\to\mathbb{R}^+_0$ is expressible as a \emph{homogeneous rational function} of the four counts, then
\begin{equation}
\mathcal{M}(\FPR,\TPR;\IR)=g(h_1(\FPR,\TPR),h_2(\FPR,\TPR;\IR))
\end{equation}
where  
\(h_1\) is the factor that survives in the balanced case
\((\IR=1)\) and  
\(h_2\) contains all explicit $\IR$-dependence.

\begin{definition}[Isocurves]A contour for value $\lambda$ is the isocurve (isoline)
\[
\bigl\{(\FPR,\TPR) : \mathcal{M}(\FPR,\TPR;\,\IR)=\lambda\bigr\}
\]
Implicit differentiation of \(\mathcal{M}(\FPR,\TPR;\,\IR)=\lambda\) gives the
contour slope:
\begin{equation}\label{eq:partial_derivatives}
\frac{d\,\TPR}{d\,\FPR}\Big|_{\lambda} =-\frac{\frac{\partial\mathcal{M}(\FPR,\TPR;\,\IR)}{\partial\FPR}} {\frac{\partial\mathcal{M}(\FPR,\TPR;\,\IR)}{\partial\TPR}}
\end{equation}
Consequently, the slope and curvature of each contour could vary systematically with the imbalance ratio. Therefore, the \textit{visual skew} of isolines is a direct indicator of how strongly $\mathcal{M}$ weights majority- vs. minority-class errors.
\end{definition}

If two contour maps of the same measure display different geometries for different $\IR$, then that measure is sensitive to
imbalance. Metrics whose contour families are invariant under $\IR$ are, by definition, imbalance-robust.

Equation \ref{eq:partial_derivatives} reveals that the geometry of every level curve could contain the factor \(\IR\); hence contour plots are guaranteed to reflect class-ratio changes whenever the metric's analytic form depends on $(P,N)$.  Inspecting the deformation of those contours as $\IR$ varies provides an immediate, graphical assessment of imbalance sensitivity. Next, several examples show how the measure can be expressed in terms of TPR and FPR, and their partial derivatives. 

\begin{example}[Accuracy] Given Accuracy isoline \( \operatorname{Acc}=\frac{TP+TN}{P+N}=\lambda \) (with imbalance ratio \( \IR \)):
\begin{equation}\label{eq:ex:accuracy_function}
\operatorname{Acc}(\FPR,\TPR;\,\IR)
\!=\!\frac{\IR\cdot\TPR + (1\!-\!\FPR)}{\IR\!+\!1}
= \!\!\!\!\!\!\!\!\!\!\!\underbrace{(1\!-\!\FPR)}_{\text{imbalance–free baseline}} \!\!\!\!\!\!\!\!\!\!\!\!
  + \!
  \left(\frac{\IR}{\IR\!+\!1}\right)\!
  \bigl[\TPR\!+\!\FPR\!-\!1)\bigr]
\end{equation}

For a free parameter \(t\in[0,1]\),
\begin{equation}\label{eq:ex:accuracy_polar}
\begin{aligned}
\mathrm{TPR}(t;\IR,\lambda) &= t \\
\mathrm{FPR}(t;\IR,\lambda) &= (1-\lambda) \;+\; \IR\,(t-\lambda)
\end{aligned}
\end{equation}

Each contour of constant accuracy $\lambda$ is a straight line in the \((\FPR,\TPR)\)--plane. The line has intercept $1\!-\!\lambda(1\!+\!\IR)$ on the \(\FPR\)-axis and slope is $\IR$, so steeper lines correspond to higher positive prevalence. When \(\IR\!=\!1\) (balanced classes) the isolines have constant slope 1, and bisect the square symmetrically for $\lambda = 0.5$ ($\FPR=1-2\lambda+ t=t=\TPR$). 

The partial derivatives, with respect to $\TPR$ and $\FPR$, are
\begin{equation}
\frac{\partial \,Acc}{\partial\, \TPR}
      =\frac{\IR}{\IR+1}
\qquad
\frac{\partial \,Acc}{\partial \,\FPR}
      =-\frac{1}{\IR+1}
\qquad
\Longrightarrow \frac{d\,\TPR}{d\,\FPR}\Big|_{\lambda} =\frac{1}{\IR}
\end{equation}

Both partial derivatives contain $\IR$, and $\frac{d\,\TPR}{d\,\FPR}\big|_{_\lambda}$ only depends on the imbalance ratio, which means that larger class--imbalance makes the contour flatter (see in the first row of Fig. \ref{fig:metrics_contours} the evolution of curves for accuracy while $\IR$ is growing $(\IR=50/50=1$, $\IR=75/25=3$, and $\IR=95/5=19$, from left to right), and above all, it is sensitive to imbalance (note that the three graphs of first row of Figure \ref{fig:metrics_contours} are quite different).
\end{example}

\begin{example}[G--mean] Given geometric mean isoline $\mathrm{G}=\lambda$,
\begin{equation}
G(\FPR,\TPR)=\sqrt{\TPR\,\cdot\TNR}=\sqrt{\TPR\,(1-\FPR)} 
\end{equation}
Interestingly, the G--mean has a different behavior. For a fixed target value \(\lambda\in[0,1]\) let \(t\in[\lambda,1]\), where \(0\le \FPR \le 1\) requires \(t\ge \lambda\):
\begin{equation}\label{eq:iso_gmean}
\begin{aligned}
\TPR(t;\IR,\lambda) &= t \\
\FPR(t;\IR,\lambda) &= 1 - \frac{\,\,\lambda^2}{t\,}
\end{aligned}
\end{equation}

Each isoline is a hyperbola‐-like curve in ($\FPR$,$\TPR$) space, symmetric with respect to  $\TPR+\FPR=1$. Because no imbalance ratio appears in Equation \ref{eq:iso_gmean}, all G--mean contours are identical for every class distribution; this analytical and visual invariance corroborates that G--mean is inherently robust to class imbalance (see in the second row of Fig. \ref{fig:metrics_contours} that the three graphs are identical, independently of $\IR$). 

The partial derivatives, with respect to $\TPR$ and $\FPR$, are
\begin{equation}
\frac{\partial \,G}{\partial\, \TPR}\!=\!\frac{1-\FPR}{2\,\sqrt{\TPR\,(\!1-\!\FPR)}}
\qquad
\frac{\partial \,G}{\partial \,\FPR}\!=\!-\frac{\TPR}{2\,\sqrt{\TPR\,(1\!-\!\FPR)}}
\,
\Longrightarrow \frac{d\,\TPR}{d\,\FPR}\Big|_\lambda \!=\!\frac{\TPR}{1\!-\!\FPR}
\end{equation}
The two partial derivatives do not contain IR, and $\frac{d\,\TPR}{d\,\FPR}\Big|_\lambda$ either, which means that the slopes are independent of the class distribution $\IR$; it is steep when $\FPR$ is large $(\FPR$$\to1-)$ and flattens as $\FPR$ approaches 0 $(\FPR$$\to0+)$.
\end{example}

\begin{example}[$\mcc$]
The case of $\mcc$ is more involved: fixing $\mcc=\lambda$ yields the quadratic relation shown in Eq. \ref{eq:ex_mcc}, from which parametric form is not informative.\begin{equation}\label{eq:ex_mcc}
(\FPR+\IR\cdot\TPR)[1-\FPR+\IR(1-\TPR)]=\frac{\IR}{\lambda^2}(\TPR-\FPR)^2
\end{equation}

Let $A = \FPR+\IR\cdot\TPR$, and $B = 1-\mathrm{FPR} + \IR\cdot(1-\mathrm{TPR})$, then
\begin{equation}
\mcc
=\sqrt{\IR}\left(
  \frac{\mathrm{TPR}-\mathrm{FPR}}{\sqrt{A\,B}}\right)
\end{equation}
The partial derivatives are
\begin{equation}\label{eq:mcc_derivatives}
\frac{\partial\operatorname{MCC}}{\partial\,\mathrm{TPR}}
   \!\!=\!\!\frac{\sqrt{IR}}
          {(AB)^{3/2}}
     \Bigl[
        AB\!
        -\!\frac{IR}{2}(\mathrm{TPR}\!-\!\mathrm{FPR})(B\!-\!A)
     \Bigr]
\quad
     \frac{\partial\operatorname{MCC}}{\partial\,\mathrm{FPR}}
   \!\!=\!\!\frac{\sqrt{IR}}
          {(AB)^{3/2}}
     \Bigl[
       -\!AB
        \!-\!\frac{1}{2}(\mathrm{TPR}\!-\!\mathrm{FPR})(B\!-\!A)
     \Bigr]
\end{equation}

Thus, from both expressions in Eq. \ref{eq:mcc_derivatives}, 
\begin{equation}\label{eq:mcc_derivative_1}
\frac{d\TPR}{d\FPR}
=\frac{2AB+(\TPR-\FPR)(B-A)}{2AB-\IR(\TPR-\FPR)(B-A)}
\end{equation}
and substituting $A$ and $B$,
\begin{equation}\label{eq:mcc_derivative_2}
\frac{d\TPR}{d\FPR}
=\frac{2\,\IR\,\TPR\,(\TPR-1)+\FPR\,(2\,\TPR-1)-\TPR}{2\,\FPR(\FPR-1)+\IR\,\FPR(2\,\TPR-1)-\IR\,\TPR}
\end{equation}

Each isoline of fixed $\mcc$ is a quadratic curve in the $(\FPR,\TPR)$-plane; it coincides with the diagonal $\TPR=\FPR$ when $\mcc=0$, is \emph{symmetric} about that diagonal only in the \emph{balanced} case $(\IR=1)$, and becomes \emph{skewed} as $\IR$ departs from $1$. The local slope on any iso--$\mcc$ contour, $\left.\frac{d\,\TPR}{d\,\FPR}\right|_{\lambda}$, depends on $\TPR$, $\FPR$ and $\IR$. It satisfies three key properties: (i) on the diagonal $\TPR=\FPR$ the slope equals $1$ (line at $45^\circ$); (ii) near the corners $(\FPR,\TPR)=(0,1)$ and $(1,0)$ it tends to $1/\IR$ (accuracy--like behavior); and (iii) its magnitude inflates where $(\FPR+\IR\,\TPR)\bigl(1-\FPR+\IR(1-\TPR)\bigr)$ is small (near the corners), revealing strong sensitivity to small errors in extreme operating regions.

Consequently, unlike accuracy (constant slope $1/\IR$) and unlike G--mean (slope independent of $\IR$), $\mcc$'s slope is location-- and imbalance--dependent: as $\IR$ grows, contours tilt and warp---flattening in some regions and steepening in others---because $\mcc$ trade--off reflects both class prevalence and the predicted positive/negative balance. This explains why, in the third row of Fig.~\ref{fig:metrics_contours}, the three $\mcc$ panels (different $\IR$) show noticeable changes in tilt and curvature, even though the diagonal at $45^\circ$ is preserved.

Therefore,  $\frac{d\,\TPR}{d\,\FPR}\Big|_\lambda$ depends on $\TPR$, $\FPR$ and $\IR$, which newly means that $\mcc$ is sensitive to imbalance. In addition, this analysis quantifies the impact of imbalance on the $\mcc$ behavior.

\end{example}

\subsection{Graphical interpretation}

Contour (isoline/level--curve) plots can show whether a measure treats false positives and false negatives symmetrically or is biased toward one type of error. This is particularly important in imbalanced datasets. In addition, by generating multiple contour plots with varying class distributions (e.g., $P=50,N=50$ vs. $P=95,N=5$), one can assess how sensitive a measure is to the underlying prevalence of classes. Measures that produce stable contour shapes under imbalance are likely to be more robust. Contour plots make it easy to identify the regions of the ($\FPR$,$\TPR$) space where a classifier would be considered acceptable, strong, random, or even detrimental. Displaying multiple contour plots side by side allows for direct comparison of measure behavior, aiding in measure selection for a specific application domain. The graphical form complements analytical expressions, providing intuitive insight into the structure and limitations of each measure.

\begin{figure}[t]
\centering
\includegraphics[width=1\textwidth]{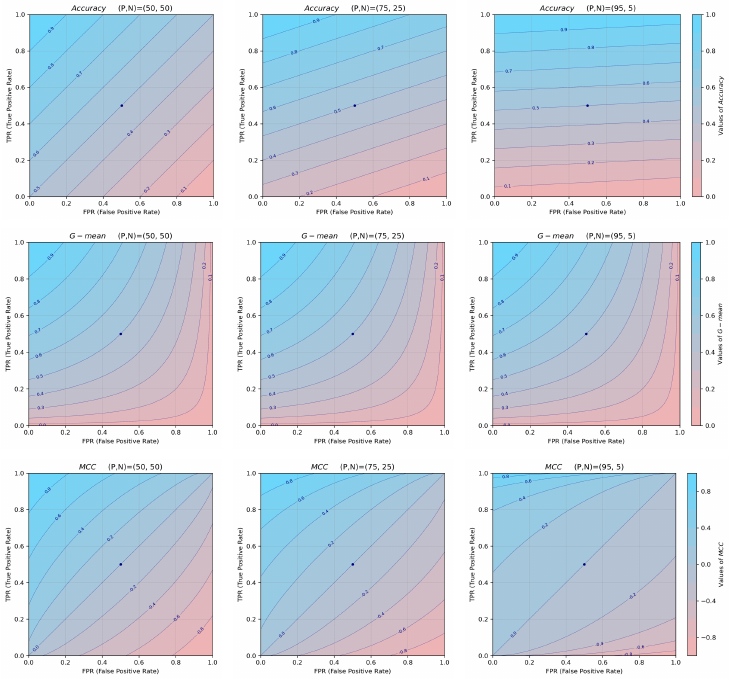}
\caption{Contour plots of several classification measures (Accuracy, G--mean and $\mcc$) under three different class balance configurations: balanced ($\IR\!=\!1$: $P\!=\!50$, $N\!=\!50$), moderately imbalanced ($\IR\!=\!3$: $P\!=\!75$, $N\!=\!25$), and highly imbalanced ($\IR\!=\!19$: $P\!=\!95$, $N\!=\!5$). Each contour is plotted in the ($\FPR$,$\TPR$) plane, showing lines of equal measure value (isolines). A dark blue point is placed in (0.5,0.5), the center of the unit square.}
\label{fig:metrics_contours}
\end{figure}

Two measures have been chosen to be compared to $\mcc$: the Finley's Index \citep{Finley1884} ---popularly known as accuracy---and the geometric mean of $\TPR$ and $\TNR$ (G--mean). These measures were selected because they are well--known and reveal very different graphical behavior.

Figure~\ref{fig:metrics_contours} presents a comparative visualization of the three chosen classification performance measures across different levels of class imbalance. Each row of the figure corresponds to a measure, while each column represents a different class distribution. The contour plots are drawn in the $(\FPR, \TPR)$ space, where $\FPR$ denotes the False Positive Rate (X--axis), and $\TPR$ the True Positive Rate (Y--axis).

\subsubsection{Accuracy}

In the top row of Fig. \ref{fig:metrics_contours}, we observe that the slopes of the accuracy contours change significantly with class imbalance. As the number of positive samples increases (from left to right), the contours shift upwards (particularly, for low $\FPR$ values), meaning that high $\FPR$ values can still yield high accuracy. For example, in the highly imbalanced case ($P=95$, $N=5$), accuracy remains above 0.8 even when the classifier generates a large number of false positives. This highlights accuracy’s well--known limitation: it is heavily biased towards the majority class, making it unreliable in imbalanced scenarios.

\subsubsection{G--mean}

The second row of Fig. \ref{fig:metrics_contours} shows the geometric mean of $\TPR$ and $\TNR$. G--mean is designed to balance sensitivity and specificity, and it does perform better than accuracy in terms of robustness. G--mean exhibits a high degree of robustness to class imbalance. It evaluates performance on both classes independently of class frequencies. This is confirmed by the visual stability of its contours across the three panels. Unlike accuracy, high $\FPR$ values can never generate acceptable G--mean values. For example, for $\FPR>0.5$, then G--mean would never exceed the value 0.7, regardless of $\TPR$.

Therefore, G--mean remains invariant under different class prevalence, making it particularly well--suited for imbalanced classification tasks where both classes are important.

\subsubsection{$\mcc$}

While $\mcc$ (bottom row in Fig. \ref{fig:metrics_contours}) is often presented as a balanced metric, it is in fact sensitive to class imbalance, especially in extreme cases. Although it accounts for all four confusion matrix components, its denominator depends on the marginal totals of each class, which introduces sensitivity to class distribution. In the figure, we observe contour distortions as imbalance increases, particularly in regions where TN or FP dominate due to low negative class size. Thus, $\mcc$ can underrepresent minority--class performance under strong imbalance. Notably, the $\mcc$ = 0 contour maintains a consistent orientation for the three $\IR$ cases, distinguishing meaningful from no--skill classification regardless of the underlying class proportions.

In general, this figure provides compelling visual evidence that $\mcc$ is not the most robust and reliable measure for evaluating binary classifiers in the presence of class imbalance. Only G--mean is prevalence--invariant by construction among the measures compared here, making it an especially robust choice in real--world tasks with skewed class distributions. Accuracy, by contrast, should be avoided in such settings due to its strong dependence on the dominant class.

The contour plot shows several important features:
\begin{itemize}
\item Imbalance effect: By plotting $\mathcal{M}(\FPR,\TPR;\IR)$ for a fixed imbalance ratio, or comparing multiple contour maps for different $\IR$, one can visually inspect: the tilt and spacing of level curves (bias toward a class), gradient magnitude (local sensitivity), and curvature sign changes (robust vs.\ vulnerable regions), thereby gaining a rigorous understanding of how $\mcc$---or any other performance measure---behaves under class imbalance without resorting merely to scalar summary statistics. This mathematical visualization complements purely numerical evaluations and offers an immediate, intuitive grasp of properties essential for reliable classifier assessment in imbalanced data contexts.

\item Bias detection: If the contour lines shift toward the $\TPR$ axis (or the $\FPR$ axis) as $\mathrm{IR}$ increases, this reveals that the measure is biased toward the majority class, disproportionately rewarding performance on the dominant class.
  
\item Robustness: Measures whose contour patterns remain symmetric and invariant under varying $\mathrm{IR}$ can be considered robust to imbalance.
  
\item Sensitivity: Measures that exhibit steep contour gradients in regions of low $\FPR$ or low $\TPR$ indicate high sensitivity to errors in the minority class---a critical property for imbalanced settings.
\end{itemize}

\section{Conclusions}\label{sec:conclusions}

Fifty years after its introduction, the Matthews Correlation Coefficient ($\mcc$) continues to be a central yet debated measure for binary classification. This work has provided a comprehensive retrospective, clarifying its historical roots, mathematical equivalence to the Pearson correlation, relation to $\chi^2$ test statistics, and its statistical significance. 

Through rigorous analysis, we have shown that while the $\mcc$ integrates all four entries of the confusion matrix and often outperforms simpler measures, it is not invariant to class imbalance and can yield misleading interpretations under extreme skew. The $\mcc$ remains a valuable measure when used with a clear understanding of its properties and limitations. 

By combining analytical derivations with graphical frameworks, most notably contour plots, we have uncovered structural properties of the $\mcc$ that scalar summaries cannot capture. These tools reveal how imbalance alters the geometry of performance measures, exposing biases, sensitivities, and robustness characteristics. Importantly, the proposed framework is not limited to the $\mcc$, but can be applied to systematically examine any classification performance metric, compare alternative metrics, and make informed choices in critical applications where reliability is paramount.

While focused on the binary case, the analytical framework introduced in this work opens promising directions for further research. Extending the $\mcc$ to multiclass classification remains an important challenge, both in terms of defining mathematically consistent generalizations and in understanding their behavior under class imbalance. Contour--based visualization and derivative--based analysis  provide a principled foundation for such extensions, as they can be adapted to higher--dimensional confusion structures and alternative performance measures. 

\bmhead{Acknowledgements}
This work was supported by Grant PID2023-152660NB-I00 funded by the Ministry of Science, Innovation and Universities, Spain.


\end{document}